\documentclass[10pt,twocolumn,letterpaper]{article}
\usepackage[pagenumbers]{cvpr} % To force page numbers, e.g. for an arXiv version
\usepackage{wrapfig}
\usepackage{enumitem}
\usepackage{graphicx}
\usepackage{amsmath}
\usepackage{booktabs}
\usepackage{enumitem}

\usepackage{bm}
\usepackage{amsmath}
\usepackage{colortbl}
\usepackage{multirow}
\usepackage{tikz}
\usepackage{wrapfig}
\usepackage{cases}

\def\checkmark{\tikz\fill[scale=0.4](0,.35) -- (.25,0) -- (1,.7) -- (.25,.15) -- cycle;} 
\newcommand{\mytilde}{\raise.17ex\hbox{$\scriptstyle\mathtt{\sim}$}}
\usepackage{algpseudocode}
\usepackage[ruled,vlined]{algorithm2e}

\newcommand{\MethodName}{Garment3DGen~}

\definecolor{cvprblue}{rgb}{0.21,0.49,0.74}
\usepackage[pagebackref,breaklinks,colorlinks,citecolor=cvprblue]{hyperref}

\def\paperID{21} % *** Enter the Paper ID here
\def\confName{3DV\xspace}
\def\confYear{2025\xspace}

\title{Garment3DGen: 3D Garment Stylization and Texture Generation}

\begin{document}
\author{Nikolaos Sarafianos, Tuur Stuyck, Xiaoyu Xiang, Yilei Li, Jovan Popovic,  Rakesh Ranjan \\
Meta Reality Labs\quad \\ 
        \normalsize{\tt\href{https://nsarafianos.github.io/garment3dgen}{nsarafianos.github.io/garment3dgen}}}
\twocolumn[{
    \renewcommand\twocolumn[1][]{#1}
    \maketitle
    \vspace{-0.3cm}
    \centering
    \includegraphics[width=0.99\textwidth]{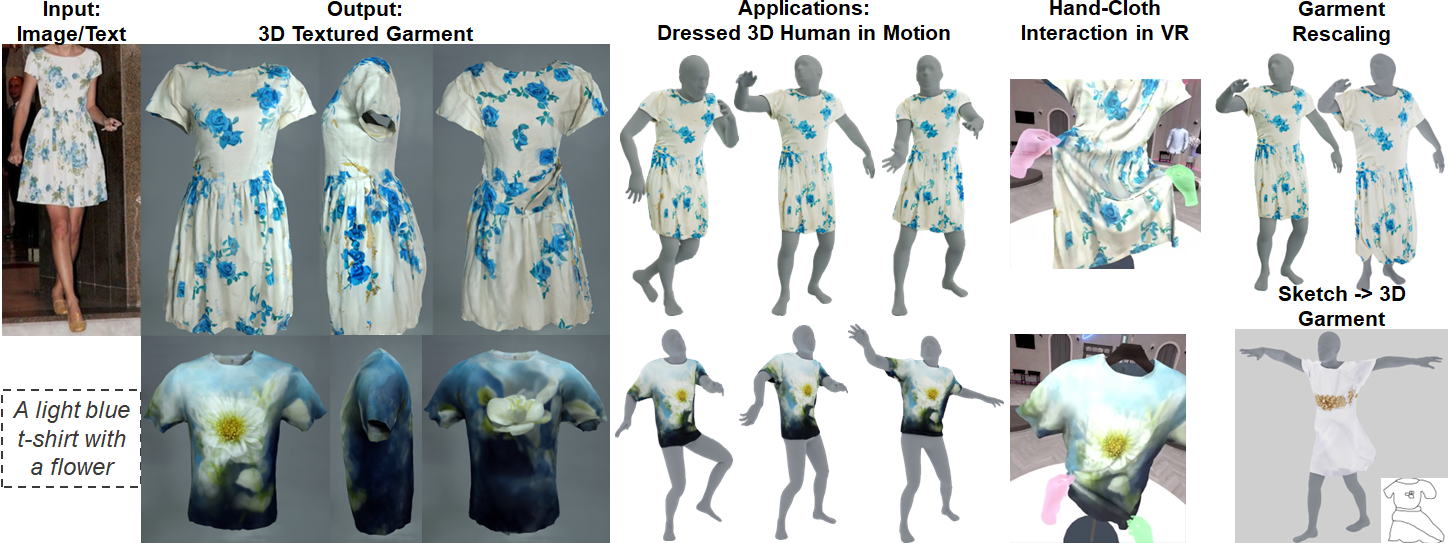}
    \captionof{figure}{\textbf{Garment3DGen}, automatically transforms a base garment mesh to simulation-ready asset directly from images or text in a frictionless manner unlocking applications such as cloth and hand-cloth interaction in a VR.}
    \vspace{0.3cm}
    \label{fig:teaser}
}]
% \vspace{0.2cm}
\begin{abstract}
We introduce \MethodName a new method to synthesize 3D garment assets from a base mesh given a single input image as guidance. 
Our proposed approach allows users to generate 3D textured clothes based on both real and synthetic images, such as those generated by text prompts. The generated assets can be directly draped and simulated on human bodies. 
We leverage the recent progress of image-to-3D diffusion methods to generate 3D garment geometries. However, since these geometries cannot be utilized directly for downstream tasks, we propose to use them as pseudo ground-truth and set up a mesh deformation optimization procedure that deforms a base template mesh to match the generated 3D target. Carefully designed losses allow the base mesh to freely deform towards the desired target, yet preserve mesh quality and topology such that they can be simulated. 
Finally, we generate high-fidelity texture maps that are globally and locally consistent and faithfully capture the input guidance, allowing us to render the generated 3D assets. 
With \MethodName users can generate the simulation-ready 3D garment of their choice without the need of artist intervention. 
We present a plethora of quantitative and qualitative comparisons on various assets and demonstrate that Garment3DGen unlocks key applications ranging from sketch-to-simulated garments or interacting with the garments in VR. Code is publicly available. 
\end{abstract}

% \vspace{-0.15cm}
\section{Introduction}

3D asset creation is the process of designing and generating geometries and materials for 3D experiences. It has direct applications across several industries such as gaming, movies, fashion as well as VR applications. 
Traditionally, simulation-ready garments are hard to obtain and are created through a laborious time-consuming process requiring specialized software \cite{CLO3D, Vstitch, style3D} relying on experienced artists. Currently, creating virtual clothing for simulation is a challenging task. Garments need to be manually designed and draped onto an underlying body. Additionally, the topology of the garment needs to take simulation requirements into consideration in order to enable stable and visually pleasing results. 
Low-friction asset creation will be the key enabler to unlock virtual applications at scale. Generative AI will be a cornerstone technology that will allow anyone, from novice users to experts, to create customized avatars and to contribute to building personalized virtual experiences. 
In addition, it will assist in the design process to facilitate faster exploration and creation of new designs. 

To tackle this task, we set out to develop a method that creates
3D garments directly from images. 
Given a base geometry mesh and a single image, \MethodName performs topology-preserving mesh-based deformations to match the image guidance and synthesizes new 3D assets on the fly. 
Our generated garments comprise of posed geometries that stylistically match the input image, and high-resolution texture maps. 
The provided image guidance can be either from the real world or synthetically generated~\cite{rombach2022high, dai2023emu} which enables us to create both real and fantastical 3D garments. 
One way to tackle this problem would be to utilize recent image-to-3D techniques~\cite{liu2023one}. Given a single image as input, such methods synthesize a specific number of views captured from pre-set viewpoints and then employ multi-view reconstruction techniques to obtain the 3D asset. 
However, the output geometries tend to be coarse and lack fine-level details due to the use of Marching Cubes to extract the output 3D geometry. 
Another limitation is that the output garments are watertight and have arbitrary scale, making it difficult to drape them on human bodies and simulate them directly. This is because manual intervention would be required to post-process the geometry. For example, creating arm, neck, and waist holes for a t-shirt geometry, as well as re-meshing to produce mesh topologies suitable for downstream applications which are sensitive to poor mesh qualities.
Alternatively, one could follow a NeRF-based approach where a handful of views of the base mesh are utilized to train a NeRF which can be stylized in an iterative manner~\cite{haque2023instruct}. %, allowing for the generation of new garments. 
However, such an approach does not guarantee multi-view consistency of the newly stylized garment because the NeRF training and stylization are happening in an iterative manner. In addition, it is a time-consuming process and the final result is not suitable for simulation.
Recent works have focused on 3D Gaussian Splatting~\cite{kerbl20233d} to generate 3D assets from image inputs. While such methods are fast and of produce high reconstruction quality, their output splats are hard to be used for any downstream task besides rendering. 
Another direction of research predicts 2D garment patterns~\cite{NeuralTailor2022,berthouzoz2013parsing,de2023perfectdart} which can be optimized using differentiable simulation~\cite{li2023diffavatar, chen2024dress}. Such approaches generate simulation-ready garments but cannot generalize to fantastical AI-generated garments and are constrained to specific garment styles. For example, when testing SewFormer~\cite{liu2023towards} on upper-body garments, it only works well on input with V-neck shirts, which is very limiting; while for other garments (long-sleeve with round neck) it fails to generate accurate patterns.

To this end, we carefully designed \MethodName to tackle each of the aforementioned challenges: i) reconstruction-based approaches output geometries that are watertight, coarse and the garments cannot be draped on human bodies ii) deformation-based approaches are under-constrained when given a single image or text prompt and their outputs do not faithfully match the provided guidance and iii) simulation-based approaches fail to generalize to new garment types. 
\MethodName produces high-quality simulation-ready stylized garment assets along with the associated textures. We approach this task from deformation-based perspective as we believe that it provides better properties and more fine-grained control for the output geometries compared to alternative NeRF-based or reconstruction-based approaches. Mesh-based deformations can preserve the mesh topology which in turn can allow for UV texture transfer, they can preserve the arm/body/head holes of the garment geometry instead of outputting watertight meshes, and can provide output meshes the triangles of which are not distorted and can be draped on human bodies and simulated. 
Our method takes as inputs a single image and a base template mesh and outputs a deformed mesh that faithfully follows the image guidance while preserving the structure and topology of the base mesh. 
Our first contribution stems from supervising the mesh deformation process directly in the 3D space which directly allows physics-inspired losses that ensure simulation-readiness instead of solely relying on image-based or embedding-based supervisions~\cite{xu2023metaclip,michel2022text2mesh,mohammad2022clip}.
In the absence of 3D ground-truth, we build upon the progress of diffusion-based multi-view consistent image generation to obtain a coarse 3D geometry that can serve as pseudo ground-truth. 
However, strictly enforcing 3D supervisions using a coarse mesh results in deformed meshes that lack fine-level details and are not simulation-ready. 
Thus, we utilize a pre-trained MetaCLIP model~\cite{xu2023metaclip} finetuned to garment data and introduce additional losses in the image as well as the embedding spaces using differentiable rendering.
Finally, we propose a carefully designed texture estimation module to predict the texture maps required to create the final 3D garment. 

We conducted a plethora of experiments that demonstrate that our method generates 3D garment assets i) directly from images allowing a frictionless experience where users indicate a requested garment by providing a reference image and quickly obtain a high-quality 3D asset without manual intervention and ii) from text describing both real and fantastical garments, and iii) even garment sketches that one can quickly draw.
Moreover, we have developed a body-garment co-optimization framework that enables us to scale and fit the garment to a parametric body model. This allows us to animate the body model and perform physics-based cloth simulation, resulting in a more accurate representation of the garment's behavior in various novel scenarios. 
In summary, our contributions are as follows: 
\begin{itemize}[leftmargin=*]
    \item We propose a new deformation-based approach for 3D geometry and texture generation for garments given a base mesh and a single image guidance as inputs. 
    % We believe that this is the first work aiming to generate textured garment assets that can be useful for downstream simulation tasks.  % \jovan{or: 'preparation'?} 
    \item We introduce geometry supervisions directly on the 3D space by generating coarse-guidance meshes from the image inputs and use them as soft constraints during the optimization. 
    % \jovan{or: 'coarse guidance meshes'?}
    In addition, we provide valuable insights on the impact of different losses that ensure that the output geometries are suitable for downstream tasks such as cloth-simulation or hand-garment interaction.
    \item We introduce a texture enhancement module that generates high-fidelity UV textures from a single image allowing us to render the output geometries. 
\end{itemize}

\vspace{-0.2cm}
\section{Related Work}
\noindent\textbf{Garment Modeling}: An important line of work is focused on designing~\cite{decaudin2006virtual,brouet2012design,wang2018SharedShapeSpace}, capturing~\cite{yu2019simulcap}, registering~\cite{halimi2022pattern}, reconstructing~\cite{qiu2023rec,corona2021smplicit,moon20223d,jiang2020bcnet,Yang2018GarmentRecovery, li2024garment}, and representing~\cite{deepcloth_su2022, liu2023gshell} clothes and their texture~\cite{3Dhumantexturesynthesis} from images or videos. BCNet~\cite{jiang2020bcnet} predicts garment vertex displacements whereas \citet{moon20223d} map 3D garments to a parametric body using dense keypoints. Both these approaches cannot generalize to loose clothes moving freely. 
\citet{guo2023diffusion} learns a shape diffusion-based prior from captured 4D data in order to enable registration of texture-less cloth, whereas \citet{lin2023leveraging} aligns the garment geometry to real world captures using a coarse-to-fine method that leverages intrinsic manifold properties with neural deformation fields.
Aiming to model clothes and discover compact ways to represent them, CaPhy~\cite{caphy_su2023} recovers a dynamic neural model of clothing similar to SNUG~\cite{santesteban2022snug} by leveraging 3D supervised training in combination with physics-based losses.
\citet{xiang2022dressing} introduced a physically-inspired appearance representation by learning view-dependent and dynamic shadowing effects. Finally, recent methods explored modeling clothes using graph neural networks~\cite{grigorev2023hood, halimi2023physgraph, pfaff2021learning}.
An alternative way to represent clothes is via sewing patterns as it ensures an efficient representation of developable~\cite{stein2018developability,rose2007developable} and manufacturable garments which can be easily modified. 
A plethora of works~\cite{qi2023personaltailor,bang2021estimating,bartle2016physics} have followed this path for garment reconstruction~\cite{liu2023towards,wang2023hisr,tiwari22posendf}, generation~\cite{shen2020gan, he2024dresscode, liu2024clothedreamer} and draping~\cite{li2023isp}.  

\noindent\textbf{Garment Deformation and Stylization}: The recent progress in language and image-to-3D models has unlocked new ways of representing and reconstructing dressed avatars~\cite{jiang2022text2human,wang2023disentangled} from text or images. Such methods typically generate multi-view consistent views~\cite{liu2023syncdreamer, weng2023consistent123, liu2023one, qian2023magic123, long2023wonder3d} given a single image or text input or directly optimize a 3D scene~\cite{poole2022dreamfusion} using a 3D scene parameterization, similar to Neural Radiance Fields~\cite{mildenhall2020nerf}. 
However such methods generate coarse, watertight meshes that in the context of garments do not have the required topology and structure to be draped on humans and simulated~\cite{stuyck2022cloth}. 
Editing and stylizing 3D surfaces has been explored for optimizing directly in 3D~\cite{sorkine2004laplacian,liu2018paparazzi} and more recently using triplanes~\cite{Fruehstueck2023VIVE3D} and text-to-mesh formulations~\cite{chen2019text2shape, mohammad2022clip}. For example, \citet{michel2022text2mesh} performs mesh stylization by predicting color and local geometric details that follow a text prompt. 
Deformation-based approaches~\cite{maesumi2023explorable,sumner2004deformation,jacobson2011bounded,baran2009semantic,wang2015linear,zhang2008deformation,gao2018automatic,groueix2019unsupervised,yifan2020neural, gao2023cloth2tex} can leverage these foundation models to enforce supervision for text and image-based stylization~\cite{decatur20233d} and manipulation~\cite{gao2023textdeformer} of 3D meshes. 
Recent work~\cite{richardson2023texture,chen2023text2tex,zeng2023paint3d,yeh2024texturedreamer} applied text-to-image generation models to create textures based on the mesh and given text/image.
Extending such techniques to clothes is a complex task as the supervision signals of a single image or text-prompt are insufficient to ensure that the deformed clothes will be simulation-ready. 

\begin{figure*}[t!]
    \centering
    \includegraphics[width=0.85\linewidth]{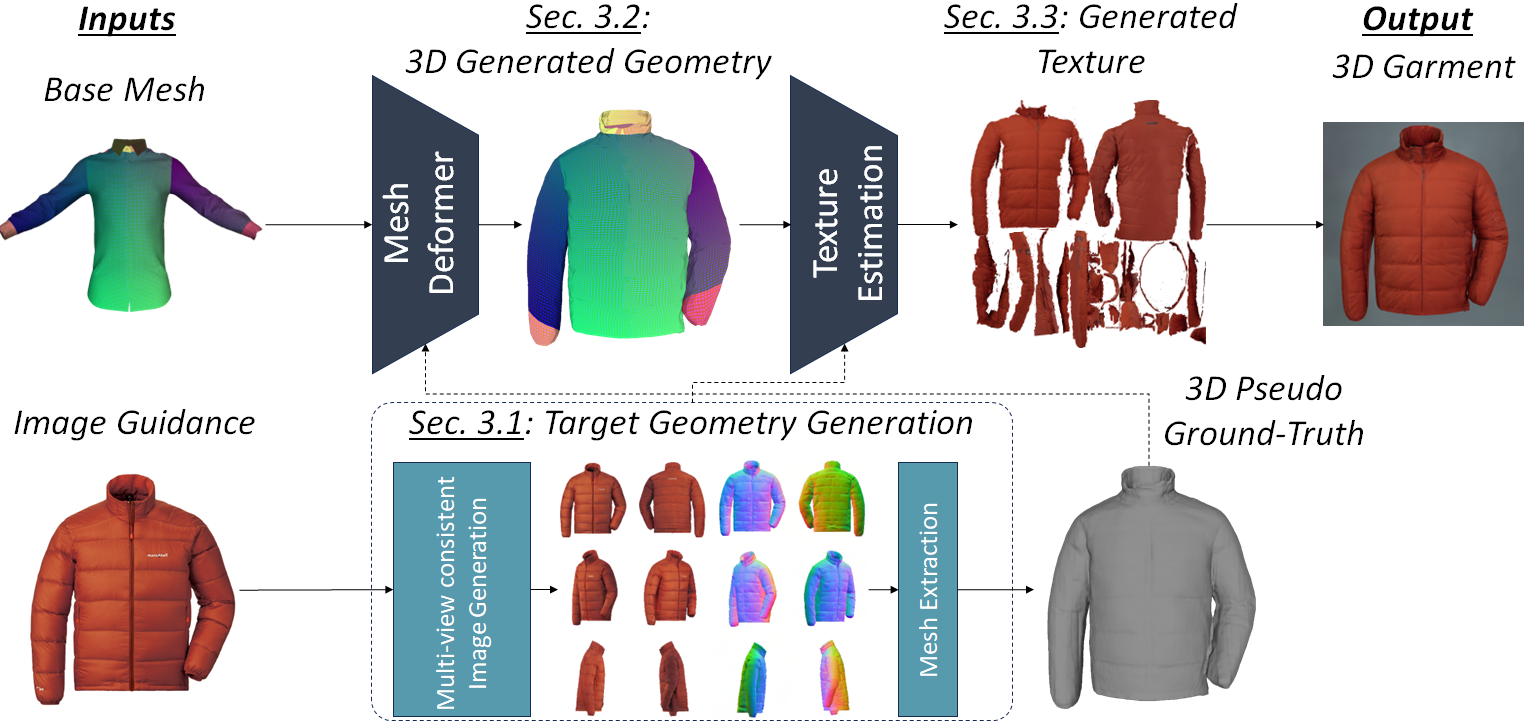}
    \caption{\textbf{Overview}: Given an input 3D base mesh and a target garment image we first generate 3D pseudo ground-truth using a diffusion-based method and utilize the output geometry as a soft supervision signal during the deformation process. Our 3D generated geometry preserves the topology and structure of the base mesh as depicted by the colors of the sleeves/collar while accurately reflecting the geometry and details of the input image. Finally, we introduce a texture-estimation module which outputs the corresponding UV texture that along with the geometry comprise our final generated 3D garment.}
    \label{fig:overview}
    \vspace{-0.3cm}
\end{figure*}

\vspace{-0.25cm}
\section{Methodology}
%\vspace{0.1cm}
Our approach takes as input a single image \(I\) and a base garment template mesh \(M_{\text{in}}\) and performs a topology-preserving deformation of the input geometry given the image guidance to obtain the target deformed mesh \(M_{\text{def}} = D(I,M_{\text{in}})\) 
% \jovan{what is 'c'?}
where \(D\) is a function that optimizes over the input mesh. An overview of our method is depicted in Fig.~\ref{fig:overview}.

\vspace{-0.2cm}
\subsection{Target Geometry Generation}\label{ssec:deformer}
We propose to leverage the recent progress of single-image-to-3D methods to obtain a coarse geometry of \(I\) denoted by \(M_{\text{guide}}(I)\) and use it as much stronger supervision both directly in the 3D space as well as in projected 2D space through differentiable rendering. 
% we choose to augment Wonder3D because of ... 
A cross-domain diffusion model~\cite{shi2023zero123plus,long2023wonder3d} is employed which synthesizes RGB and normal images from six views given the input image \(I\) captured from the same predefined viewpoint. A multi-view 3D reconstruction algorithm based on the LRM architecture~\cite{xu2024instantmesh} is then utilized that, given the generated views, it outputs a watertight, relatively coarse geometry \(M_{\text{guide}}(I)\) of the garment in the input image. 
Such meshes cannot serve as the final simulation-ready result due to its poor mesh quality which is due to Marching Cubes~\cite{lorensen1987marching} (or FlexiCubes~\cite{shen2023flexible}) or potential inaccuracies of the multi-view generation. Additionally, the fact that it is watertight prevents us from draping the garment on a body (\eg missing armholes). Nonetheless, it provides useful information to serve as a pseudo ground-truth that serves as a reference to deform \(M_{\text{in}}\) towards. Hence, we update the optimization function as follows \(M_{\text{def}} = D(I, M_{\text{in}}, M_{\text{guide}}(I))\).
The alternative approach would be to rely on the input image \(I\) as the sole supervision for mesh deformation which would result in a severely under-constrained optimization with low-quality output meshes that are uncanny, over-deformed, and they fail to capture the subtle details of the image guidance. 
For example, starting from a template T-shirt geometry with guidance of the image of an armor, one could extract CLIP embeddings for both \(I\) and the renders of \(M_{\text{def}}\) following a similar approach to TextDeformer~\cite{gao2023textdeformer}. 
By enforcing supervisions on the embeddings, the goal is obtain an output mesh that would resemble the requested armor. In practice, the supervision from the embedding of a single exemplar image is not strong enough to produce high quality output.

\vspace{-0.2cm}
\subsection{Topology-Preserving Deformations}\label{ssec:geometrygen}
In order to preserve the structure and topology of the input base mesh while enabling image-base stylizations, we propose an approach which deforms \(M_{\text{in}}\). In contrast, generating novel geometry using reconstruction-based methods would be hard to use directly in downstream tasks. Hence, inspired by Neural Jacobian Fields~\cite{aigerman2022neural}, we parameterize \(M_{\text{in}}\) using a set of per-triangle Jacobians which define a deformation. 
Following the same formulation, for simplicity we represent per-triangle Jacobians as matrices \(J_i \in R^{3x3}\) and solve a Poisson optimization problem to obtain the deformation map \(\Phi^*\) as the mapping with Jacobian matrices for each triangle that are closest to \(J_i\). More formally this is represented as: %for every face
\setlength{\abovedisplayskip}{3pt}
\setlength{\belowdisplayskip}{3pt}
\begin{equation}
    \Phi^* = \min_{\Phi} \sum |t_i| \lVert \Phi \nabla_i^T - J_i \rVert ^2,
\end{equation}
where \(\nabla(\Phi)\) denotes the Jacobian of \(\Phi\) at triangle \(t_i\), with \(|t_i|\) being the area of the triangle. 
We optimize the deformation mapping \(\Phi\) indirectly by optimizing the matrices \(J_i\) which define \(\Phi^*\). 
These Jacobians are initialized to identity matrices. 
With the Jacobian representation at hand, we optimize over the triangles of \(M_{\text{in}}\) by introducing a several losses each addressing a specific issue. 

\noindent\textbf{3D Supervisions}: We employ the one-directional Chamfer Distance (CD) loss to evaluate the similarity between sets of points \(p_{\text{def}} \in S_{\text{def}}\) and \(p_{\text{I}} \in S_{\text{I}}\) sampled randomly in each iteration from \(M_{\text{def}}\) and \(M_{\text{guide}}(I)\). This is defined as:
\setlength{\abovedisplayskip}{3pt}
\setlength{\belowdisplayskip}{3pt}
\begin{equation}
    L_\text{CD} = \frac{1}{|S_{\text{def}}|}\sum_{p_{\text{def}} \in S_{\text{def}}} \min_{p_{\text{I}} \in S_{\text{I}}} \lVert p_{\text{def}} - p_\text{I}\rVert_2^2.
\end{equation}

\noindent\textbf{Regularizations}: We introduce several regularizations on the deformed 3D mesh to ensure that it maintains key properties. First, we introduce Laplacian smoothing~\cite{field1988laplacian} denoted by \(L_\text{Lap}\) to redistribute vertex positions based on the average positions of neighboring vertices. This smoothing process helps to reduce irregularities and improves the overall mesh shape. 
To produce simulation-ready meshes we penalize very small surface area triangles denoted by \(L_\text{triag}\) as that would result in meshes that are difficult to simulate. We do this by regularizing the edge length and by minimizing the inverse of the squared sum of the triangle areas. 
Note that there is a trade-off between how much a mesh can freely deform, (\eg a shirt becoming a spiky medieval armor), and how much regularization it requires such that the garment can be placed on a parametric body and simulated. 

\noindent\textbf{2D Supervisions}: We utilize a rasterization-based differentiable renderer~\cite{laine2020modular} denoted by \(R\) and pass both the deformed mesh \(M_{\text{def}}\) in each iteration and target pseudo ground-truth mesh \(M_{\text{guide}}(I)\) to obtain \(K\) image renders \({I_{\text{def}}}_i = R(M_{\text{def}}, C_i), \;\;\; i=1\dots K\) from randomly sampled camera views \(C_i\).
With \({I_\text{I}}_i\) computed in a similar fashion for \(M_{\text{guide}}(I)\) we employ the L1 loss between the deformed and target renders:  
\begin{equation}
    L_\text{2D} = \frac{1}{K}\sum_{i=1}^K \lVert {I_{\text{def}}}_i - {I_\text{I}}_i \rvert.
\end{equation}
This supervision in the 2D space captures well the silhouette of the garment from multiple views as well as its fine-level details thereby enforcing the deformed mesh to not deviate far from the target along each step of the optimization. 

\noindent\textbf{Embedding Supervisions}: We observe that passing garment images through a pre-trained CLIP model results in deformed output garments that are overly distorted, uncanny and fail to capture the fine-level details provided in the input images. This is due to the weak supervision signal contained in these embeddings. 
Likewise, this holds true for other mesh classes that are not well represented in the data on which CLIP was trained on, such as human geometries for example. This is because CLIP fails to capture the subtle differences between garments, their properties and materials. 
To overcome this limitation, we propose to use garment-specific embeddings obtained from a CLIP model fine-tuned on fashion data named FashionCLIP~\cite{Chia2022} denoted by FCLIP. The latent space for this model is better tuned for fashion concepts and as a result, the embeddings provide a stronger guidance for the deformations. We represent this loss as: %
\begin{equation}
    L_\text{E} =  \frac{1}{K}\sum_{i=1}^K \text{Cos} \left( \text{FCLIP} \big( {I_{\text{def}}}_i \big), \text{FCLIP} \big( {I_\text{I}}_i \big) \right),
\end{equation}
where \textit{CosSim} is the cosine similarity. 
This embedding loss acts as a soft supervision signal between the embeddings of the deformed mesh \(M_{\text{def}}\) and those of the pseudo-ground-truth \(M_{\text{guide}}(I)\). 
This behavior is desired since we aim to benefit from the embedding representations of the CLIP model and sufficiently deform the base mesh towards the target without capturing all of its shortcomings. For example we need to preserve the arm/head holes of the base geometry while the target mesh is watertight, or capture the fine-level details depicted in the input image that \(M_{\text{guide}}(I)\) might have failed to represent well.
In summary, the total loss \(L_\text{T}\) is defined as follows where \(w_{*}\) is the corresponding weight for each loss: 
\setlength{\abovedisplayskip}{3pt}
\setlength{\belowdisplayskip}{3pt}
\small
\begin{equation}
    L_\text{T} = w_\text{CD}L_\text{CD} + w_\text{Lap}L_\text{Lap} + w_\text{triag}L_\text{triag} + w_\text{2D}L_\text{2D} + w_\text{E}L_\text{E}, 
\end{equation}
\normalsize
where the corresponding weights balance between utilizing the 3D pseudo ground-truth with the CD loss while capturing the finer details through the embedding and 2D losses.

\vspace{-0.1cm}
\subsection{Texture Estimation}\label{ssec:texture}
\vspace{-0.1cm}
Given the untextured deformed 3D geometry $M_{\text{def}}$, we generate high-fidelity textures that match the input image. 
We leverage a 2D text-to-image generation model to create high-quality, high-resolution textures with vivid details given a text prompt. When starting from images where text is absent, a text prompt is obtained using an image caption model~\cite{li2023blip}. 
Our pipeline consists of the following steps: generating images of the given fashion asset from multiple views and backprojecting these images onto the mesh surface to create UV texture $T\in \mathbb{R}^{H\times W \times C}$. There are two challenges in adapting a 2D generation model to 3D objects: establishing geometry-texture correspondence and ensuring multi-view consistency.

\noindent \textbf{Shape-Aware Generation}: To ensure the texture faithfully reflects the underlying shapes, we propose using a depth-aware text-to-image generation model. Given a set of camera poses $\mathcal{C}=\{C_i\}_{i=1}^{n}$,
we render the depth $D_i$ of each view from $M_{\text{def}}$, and sample the appearance image $I_i$ of view $C_i$ conditioned on $D_i$ and the text prompt using the text-to-image model.

\noindent \textbf{Multi-view Consistency}: Directly conducting view-by-view image synthesis cannot guarantee that the generated views are consistent with each other. To solve this problem, we first synthesize the front and back views simultaneously to implicitly enforce global consistency. 
% \jovan{this is because front and back views don't overlap?}->true!
Then, we leverage the fact that the geometry of fashion assets is mostly flat, and conduct depth-aware inpainting for the remaining views to ensure the newly generated textures are locally consistent. 
To handle occluded areas, we design an automatic view selection algorithm to inpaint the textures in a coarse-to-fine manner: from the remaining unpainted area, we select the view with the most unfilled pixels to generate the textures iteratively from large regions to small pieces.
% \jovan{can this be written up more formally in pseudocode?}->can be put into supp due to space limit

\noindent \textbf{Texture Enhancement}: The above texture generation pipeline can also be applied to texture refinement: given a low-quality, low-resolution initial texture $T^{LQ}$, we leverage the 2D appearance priors to further enhance the details. We adopt SDEdit~\cite{meng2021sdedit}, which perturbs the above sampling procedure with Gaussian noise and progressively denoises by simulating the reverse stochastic differential equations. As a result, the low-quality $I_i^{LQ}$ is projected onto the manifold of realistic images, yielding $I_i^{HQ}$. By backprojecting these images, we acquire a high-quality texture $T^{HQ}$.

\noindent Our texture estimation module has the following improvements compared with the previous works: 
\begin{itemize}[leftmargin=*]
    \item \emph{Global consistency}: by generating the front and back views together, we enforce better global consistency. 
    \item \emph{Geometry-aware texture inpainting}: Past works tweaked the depth-to-image generation model with masked generation methods (TEXTure~\cite{richardson2023texture} uses Blended Diffusion, and Text2Tex~\cite{chen2023text2tex} uses RePaint). We directly use multi-ControlNet, enabling a better balance between the depth and mask controls, which produces better local content consistency. 
    \item \emph{Backprojection}: we implement a faster backprojection method using PyTorch $grid\_sample$, which has the lowest latency compared to all known methods.
    \item \emph{Speed}: Our texture generation is significantly faster than: TEXTure: 2.5 mins, Text2Tex:12mins, Ours: 4.56s
\end{itemize}
We refer the reader to the supplementary for an in-depth discussion of trade-offs.

\begin{figure*}[t]
    \centering
    \includegraphics[width=0.99\linewidth]{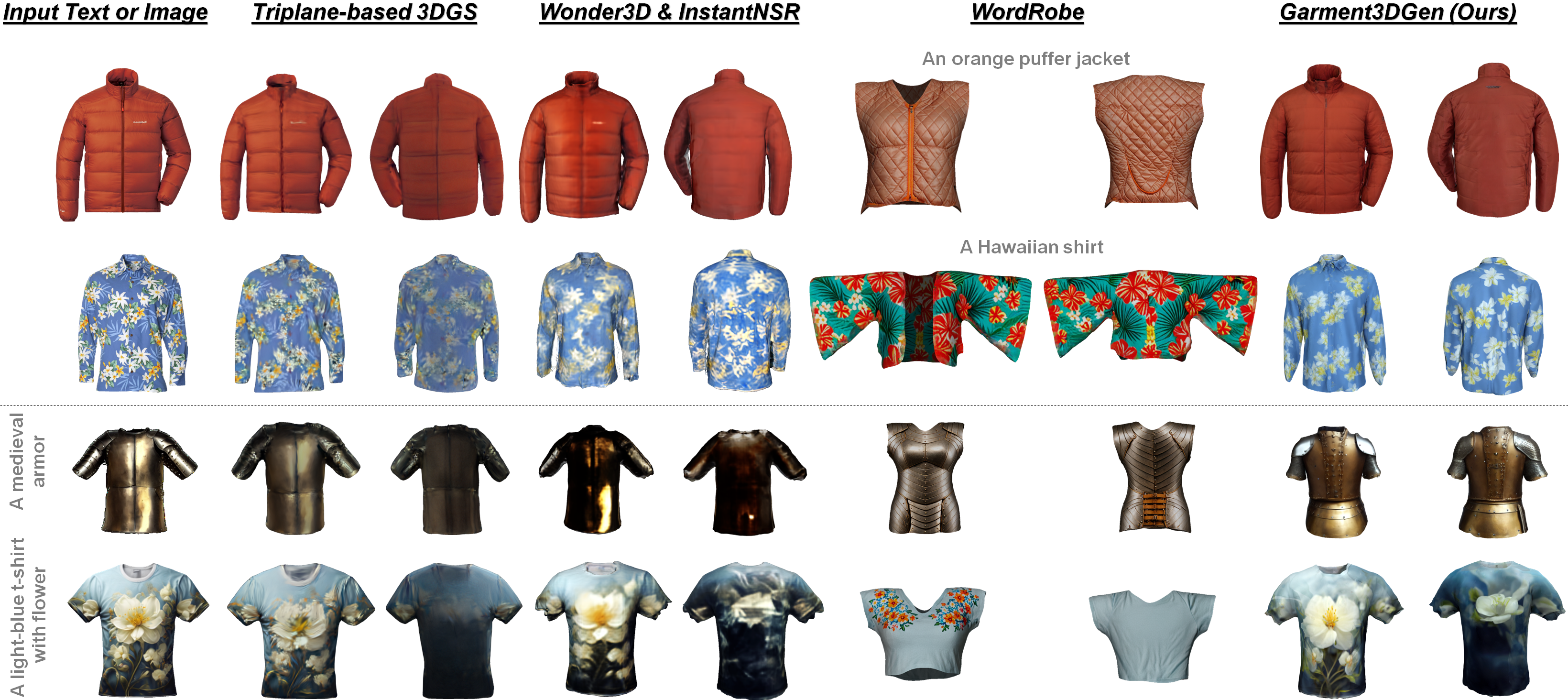}
    \caption{\textbf{Qualitative Comparisons}: We demonstrate several mesh generation methods given an image (top) or text (bottom) and show front and back views of each reconstruction. The 3D Gaussian Splatting~\cite{3dgs_triplane} method generates distorted frontal colors and dark or blurry back colors while its geometry is not suitable for downstream tasks such as simulation. The second reconstruction approach~\cite{long2023wonder3d,instant_nsr} generates watertight meshes with coarse geometric details and blurry colors. WordRobe\cite{srivastava2024wordrobe} which is purely text-based, generates simulation-ready garments but they deviate far from the text prompt (\eg, the puffer jacket and t-shirt are not faithful to the prompt). Our proposed approach outputs 3D geometries that are geometrically correct with fine-level texture details that prior works fail to generate.}
    \label{fig:qualfull1}
\end{figure*}

\vspace{-0.2cm}
\section{Experiments}
%\vspace{-0.15cm}
\noindent\textbf{Data}: We use the publicly available DiffAvatar~\cite{li2023diffavatar} dataset which comprises of artist-created garment templates covering several cloth categories (\eg T-shirt, shirt, tank-top, dress, etc). 
To demonstrate garment generation using image guidance, we collect a variety of real images with garments in different poses, different textures, including garment types that are not covered by our dataset. In addition to real images, we include AI generated images using a text prompt of both realistic and fantastical garments.

\noindent\textbf{Metrics}: Quantitatively evaluating our results is a challenging task in the absence of 2D or 3D ground-truth for what are we trying to accomplish. 
However we can evaluate how consistent the geometries are to the input image and hence we render untextured outputs for all methods from 36 views and compute their perceptual scores using the LPIPS metric~\cite{zhang2018unreasonable} as well as their image-based CLIP similarity score using the cosine distance between their embeddings. 

\noindent\textbf{Baselines}: We evaluate our approach against: i) TextDeformer~\cite{gao2023textdeformer} which deforms input meshes based on text, ii) ImageDeformer - a variation of TextDeformer we developed where the input text is replaced with an image, iii) Wonder3D~\cite{long2023wonder3d} and iv) Zero123++~\cite{shi2023zero123plus} both of which generate 3D geometries given a single image using 2D diffusion models, v) ZeroShape~\cite{huang2023zeroshape} which performs zero-shot reconstruction and vi) a triplane-based 3DGS approach~\cite{3dgs_triplane}. The first two works also take as input a base mesh whereas the latter four reconstruct the final result given a single image as input. 
Finally, we perform qualitative comparisons with WordRobe~\cite{srivastava2024wordrobe} which is a recent top-performing method among some concurrent text-based garment generation methods~\cite{he2024dresscode, liu2024clothedreamer, li2024garmentdreamer} and a pattern-based approach termed SewFormer~\cite{liu2023towards} (see supplementary).

\begin{figure*}[t]
    \centering
    \includegraphics[width=0.97\linewidth]{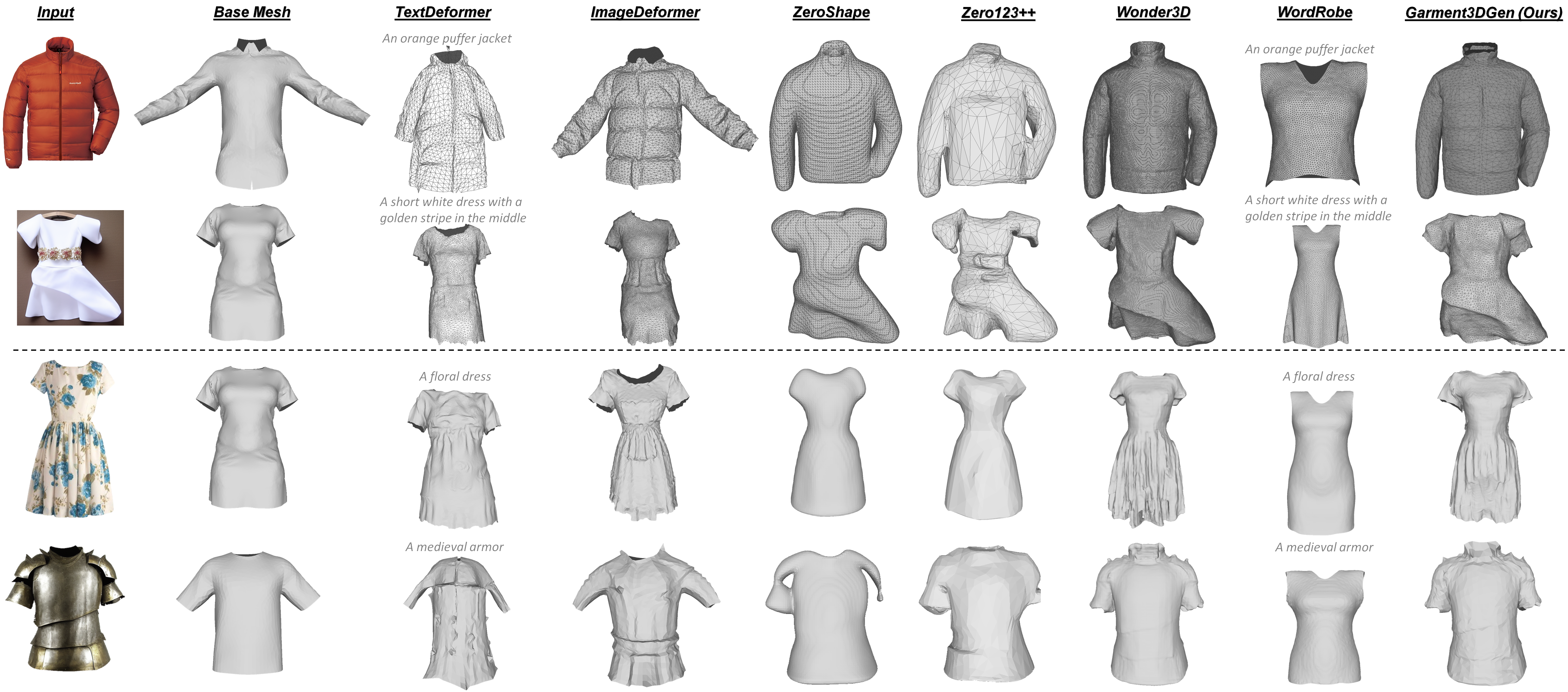}
    \caption{\textbf{Mesh Quality (Top) and Geometry Comparisons (Bottom)}: We showcase the wireframes of all approaches. Our method stands out as the only one that produces geometries that adhere to the input image while maintaining good mesh quality and incorporating necessary holes for physics-based simulation tasks. At the bottom we showcase the output geometry of various techniques to highlight that our approach captures fine geometric details without geometric artifacts (Wonder3D). Note that several methods produce smooth meshes where details are incorrectly captured in the texture map instead which results in lower quality visual results.}
    \label{fig:quality}
    \vspace{-0.2cm}
\end{figure*}

\vspace{-0.15cm}
\subsection{Image-to-3D Garments}
\begin{figure*}[t]
    \centering
    \includegraphics[width=0.95\linewidth]{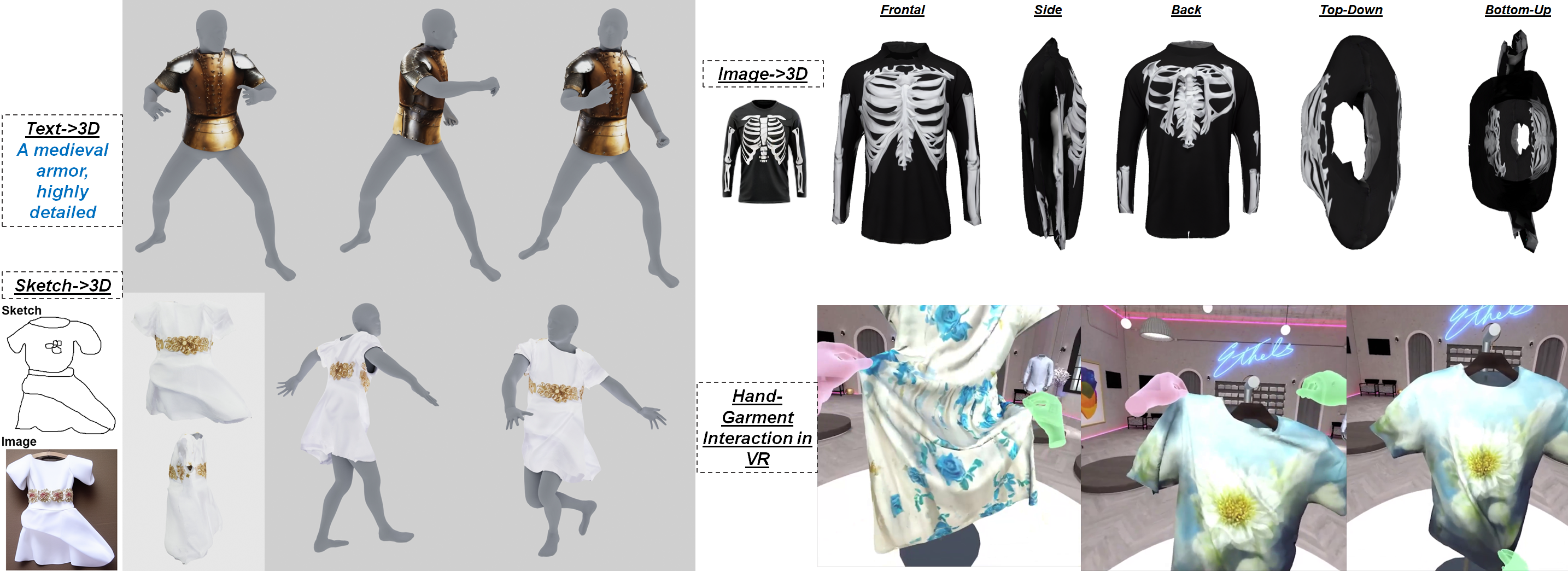}
    \caption{\textbf{Applications}: \MethodName generates textured 3D garments from images, text prompts, simple sketches, that can be fitted to human bodies and drive them with physics-based cloth simulation or even enable interaction between hands and garments in VR.}
    \label{fig:qualfull2}
    \vspace{-0.3cm}
\end{figure*}

%\vspace{-0.15cm}
Given an image prompt and a base mesh, \MethodName generates textured 3D garments matching the input image guidance. We present a plethora of image-to-3D results in Fig.~\ref{fig:teaser}, Fig.~\ref{fig:qualfull1} and Fig.~\ref{fig:qualfull2} and showcase that the generated 3D assets can be of various topologies, respect the pose, shape and texture of the input image. Garments can be both real or fantastical. They contain high-quality texture maps which exhibits fine-level details such as cloth wrinkles and folds and preserves the topology and structure of the input (\eg, head, bottom and armholes are respected such that the garments can be draped on a body). 
Additionally, we quantitatively evaluate \MethodName against several baseline methods and report our results in Table~\ref{tab:quantitative}. Our approach outperforms all methods in terms of both embedding as well as perceptual similarity with the input image guidance while it is the only approach that generates textured geometries that can be used for downstream tasks. Deformation-based approaches (rows 1-2) preserve the topology and the holes of the garment but lack strong supervision signals to generate outputs that match the input image. Reconstruction-based approaches (rows 3-6) are good at preserving and reconstructing what is visible in the input image but generate unusable geometries (or splats) and their color estimates for the non-visible regions are usually blurry or single colored (as shown in Fig.~\ref{fig:qualfull1}). 

% \vspace{-0.15cm}
\subsection{Text-to-3D Garments}
\vspace{-0.15cm}
We easily extend our work to enable textured garment generation using textual prompt inputs. Unlike TextDeformer~\cite{gao2023textdeformer} or WordRobe~\cite{srivastava2024wordrobe} which utilize text directly, we opt for a text-to-image diffusion model as an initial step. We do this because image-based supervisions provides a stronger guidance for our mesh deformation process which produces better results. 
This allows the user to easily iterate with several text-prompts to generate the image of their desired garment . This is in stark contrast to the slow iteration time of providing a text-prompt and waiting for the 3D asset creation process to finish to see if the result matches their intent. 
In Fig.~\ref{fig:qualfull1} (bottom) we present generated garments using only the provided textual prompt as guidance.

\begin{table}[t]
    \centering
    \setlength{\tabcolsep}{0.75mm}
    \renewcommand{\arraystretch}{1.2}
    \caption{\textbf{Quantitative Comparisons}: Our approach outperforms deformation-based (rows 1,2) and reconstruction-based (rows 3-6) methods across both metrics while generating textured geometries that can be used for downstream tasks which is not the case for any of the prior methods.}
    \resizebox{\columnwidth}{!}{
    \begin{tabular}{lccccc}
        \toprule
        Method & CLIP-Sim$\uparrow$ & LPIPS $\downarrow$ & Faithful to Image & Colored Output & Head/Arm Holes\\ 
        \midrule
        TextDeformer~\cite{gao2023textdeformer} & 0.51 & 0.42 & & & \checkmark\\% &  \\
        ImageDeformer~\cite{gao2023textdeformer} & 0.54 & 0.41 & & & \checkmark \\%&  \\
        \hline
        Wonder3D~\cite{long2023wonder3d}  & 0.56 & 0.41 & \checkmark & \checkmark & \\%& \\ %\cite{long2023wonder3d,instant_nsr}
        Zero123++~\cite{shi2023zero123plus}  & 0.52 & 0.42 & \checkmark & \checkmark & \\%& \\
        ZeroShape~\cite{huang2023zeroshape}  & 0.48 & 0.46 & \checkmark & & \\%& \\
        T-3DGS~\cite{3dgs_triplane} & 0.57 & 0.41 & \checkmark & \checkmark &  \\%& \\ %~\cite{3dgs_triplane}
        \hline
        \textbf{\MethodName} & \textbf{0.59} & \textbf{0.39} & \checkmark & \checkmark &  \checkmark \\%&  \checkmark \\
      \bottomrule
    \end{tabular}}
    \label{tab:quantitative}
    \vspace{-0.3cm}
\end{table}

\vspace{-0.1cm}
\subsection{Ablation Studies}
\vspace{-0.05cm}
We conducted a variety of ablation studies to assess the impact of the key components of our proposed approach the figures of which are provided in the supplementary material. 
We start with the off-the-shelf TextDeformer which takes a text prompt and a base-mesh and deforms this to match the target text. Text prompts are not ideal to capture the fine-level details of a garment as there can be many ``medieval armors''. In addition, a pretrained CLIP model is not capable of capturing the subtle differences between a ``jacket'' and a ``puffer jacket''. 
To overcome this limitation we adapt TextDeformer to take image inputs as guidance (ImageDeformer) and observe that the deformed geometries are improved. Nonetheless, they still fail to capture the details of the image. By swapping the original CLIP model and introducing a model that is fine-tuned on fashion data we observe that details are better preserved across garments. 
Noting that image-based reconstruction methods can accurately capture geometry but produce coarse and watertight meshes that are unsuitable for subsequent tasks, we utilize these meshes as pseudo ground-truth for our proposed approach.
Our \MethodName results in garments that faithfully follow the image guidance while containing the wrinkles and fine details. 
However the quality of the output geometry is not always ideal for physics-based downstream tasks because they produce poorly conditioned triangles which will result in instabilities when simulated in addition to poorly tessellated geometry which will result in unnatural fabric behavior. Because of this, we introduced additional 3D supervisions that preserve a better mesh quality, see Fig.~\ref{fig:quality}. 
All prior works either do not follow the input image guidance or generate low-quality geometries that cannot be directly used for downstream tasks. Finally we conducted ablation studies to assess the impact of the pre-trained CLIP model on garment data, the impact of regularizations as well as the texture enhancement module and refer the interested reader in the supplementary material for additional information. 

\vspace{-0.15cm}
\subsection{Applications}

\noindent\textbf{Physics-based Cloth Simulation} 
We consider the generated 3D shape as rest shape for our simulations and incorporate a zero rest-angle dihedral energy to model the out-of-plane banding of the fabric. 
Fantastical garments such as armors are not well modeled using a zero rest-angle as they would lose their distinct shape during simulation and wrinkle unnaturally. 
To obtain visually pleasing results, we take the rest angle to be the 3D generated mesh one which allows it to maintain its shape throughout the simulation. %~\cite{grinspun2003discrete}.
Generated garments can further be manipulated to achieve distinct looks through garment resizing. 

\noindent\textbf{Hand-Cloth Interaction in VR} Our garments are suitable for real-time simulations with hand-interaction using modern VR headsets.
The rightmost part of Fig.~\ref{fig:teaser} and Fig.~\ref{fig:qualfull2} show a user interacting with the garments through real-time simulation with integrated hand-tracked interactions.

\noindent\textbf{Sketch to Garment}:  
Given a rough dress sketch, we generate a realistic image using ControlNet \cite{zhang2023adding} which then serves as the input to our method to generate the corresponding 3D asset. In Fig.~\ref{fig:qualfull2} we show simulation results of this automatically generated dress.

\vspace{-0.15cm}
\subsection{Discussion}
\noindent\textbf{Simulation-ready Garments}: 
Our method stands out as the favorable choice for producing garments that are faithful to the prompt with high-quality meshes that can be seamlessly integrated into simulation frameworks as demonstrated by a qualitative and quantitative analysis. Although it might be possible to produce simulation-ready garments using alternative methods, this would require remeshing and potentially several additional manual post-processing steps, such as creating arm holes.
Moreover, our use of template meshes ensures a superior UV layout, which is challenging to attain automatically.

\noindent\textbf{Runtime}: \MethodName takes \mytilde 5 minutes on a single H100 which is 3x faster than~\cite{michel2022text2mesh,gao2023textdeformer,li2024garmentdreamer} yet slower than 3DGS-based methods that take a few seconds. While the latter render well, how to accurately re-mesh them and fit them onto bodies is an open research problem. 

\noindent\textbf{Limitations}: \MethodName handles a variety of garment types both realistic and fantastical. Due to the requirement of a template mesh, there is a limitation on what garments can be generated whilst still providing distortion-free meshes. 
This can be mitigated by providing a more diverse template library. Our estimated textures, while faithful to the image, sometimes do not fully preserve fine-level details. We plan to address this by tuning the texture enhancement module to be conditioned on the reference image across all views while maintaining its multi-view color consistency properties. Finally, it is worth noting that the closer the input template mesh is to the target garment, the easier the deformation task becomes. We currently select the closest template manually but performing automatic retrieval based on the image could be tackled in future work.

\vspace{-0.2cm}
\section{Conclusion} 
\vspace{-0.15cm}
We proposed a new approach to generating high-quality garment assets that can be directly used for downstream applications which require good mesh quality. We introduced a deformation-based approach that takes a base mesh as input along with an image guidance and outputs textured geometry that faithfully matches the input image while preserving the structure and topology of the input mesh. 
Our key contributions stem from utilizing novel diffusion-based generative models to synthesize 3D pseudo ground-truth that can be used as a soft supervision signal along with additional regularizations, a texture enhancement module that generates high-fidelity texture maps and a body-cloth optimization framework that fits the generated 3D garments to parametric bodies. 
Our approach clearly outperforms prior work. Finally, we showcased physics-based cloth simulation, hand-garment interaction in a VR environment.

\section*{Acknowledgments}
We thank Michelle Guo, Will Gao and Astitva Srivastava for their help and support on running baseline comparisons.
{
    \small
    \bibliographystyle{ieeenat_fullname}
    \bibliography{References}
}
\clearpage

\section*{Supplementary  Material}

We refer the interested reader to the supplemental video where we provide a wide variety of results ranging from image/text to 3D textured garments as well as applications of our method in downstream tasks such as physics-based cloth simulation, hand-garment interaction in VR using a headset and sketch to 3D garment reconstruction. Below we provide some additional details regarding the implementation of our key components as well as some additional ablation studies to showcase the impact of our design decisions.

\subsection*{\MethodName General Details }
We believe that our approach provides three key insights that will be valuable to the community: 
\begin{enumerate}
    \item Mesh-based deformations provide the right properties to generate (or stylize) new garments that we can utilize for downstream tasks other than rendering. 
    \item A text-prompt or a single image alone cannot provide enough guidance to generate the desired garment exactly the way a user might want it. This is evident from the results of WordRobe~\cite{srivastava2024wordrobe} which despite its mesh-quality the generated garments do not follow the provided text prompt.
    \item 3D supervisions, if done right, can provide strong enough supervision signal in order to generate the desired garments with the proper topology and structure.
\end{enumerate}
Our approach builds upon these insights and introduces a novel yet simple solution to generate high-quality, physically plausible garments. 
As input to the method, we require only a single garment image (or alternatively, a text prompt that can generate this image using a text to image model) and a base garment template mesh. The input image needs to contain a single piece of clothing, captured from a semi-frontal viewpoint with its pose being as occlusion free as possible. A person can be wearing this garment or there might be more than one piece of clothing in the image in which case we perform semantic segmentation (using SAM) to obtain the garment. 
The template mesh is not required to be similar to the image guidance. For example, we demonstrate results where our method can go from a shirt to a puffer jacket, from a tank-top to a dress or even a T-shirt to a fantastical sea armor. 
Note that the closer the base mesh is to the target geometry, the easier the task is. For example, starting from a dress mesh to go to a shirt is a difficult task while starting from something closer to the target simplifies this problem. 

\noindent\textbf{Automatic View Selection}: 
The goal of this algorithm is to automatically select the least-painted view and paint it. In this way, we can solve the 3D texture generation problem in a coarse-to-fine manner, and ensure the overall consistency. Alg.~\ref{alg:1_auto_view_selection} provides a detailed description of the automatic view selection algorithm: given the input UV texture $T$ with painted front and back views, there could be $N$ candidate views. We maintain a binary mask $T^B$ that marks the painted pixel as 1, and unpainted pixel as 0. We can select the view with the most unfilled pixels as the next view to generate the appearance, and update the binary mask $T^B$. This process is repeated iteratively until most of the pixels are painted, or reaching a certain iteration number.
\begin{algorithm}[t]
\SetAlgoLined
\textbf{Input:} an input mesh $M_{\text{def}}$ with UV texture $T$ with front and back views painted, a binary mask $T^B$ marking the painted pixels of $T$, and $N$ uniformly distributed candidate views $\{C_i\}_{i=1}^{N}$\;
 \For{number of iterations}{
 Calculate the binary mask $T^B_i$ for each view $i$ from $T^B$: $\{T_i^B\}_{i=1}^N$\;
 Select the least painted view $C_j$: $j \leftarrow \arg\min_{i=1}^{N} \sum T_i^B$ \;
 Generate the appearance image $I_i$ and update $T^B$;
 }
 \caption{Automatic View Selection}
 \label{alg:1_auto_view_selection}
\end{algorithm}

\begin{table*}[t]
    \centering
    \setlength{\tabcolsep}{0.75mm}
    \renewcommand{\arraystretch}{1.2}
    \caption{Comparisons of different texture estimation methods. The runtime is measured on a single NVIDIA H100 GPU.}
    \resizebox{\linewidth}{!}{
    \begin{tabular}{lccr}
        \toprule
        Method & \textbf{Pros} & \textbf{Cons} & \textbf{Runtime}\\
        \midrule
        Mesh2Tex~\cite{bokhovkin2023mesh2tex} & Infinite resolution \& Global consistency & One model per class \& No fine details  & 8mins\\
        TEXTure~\cite{richardson2023texture} & Shape-aware \& Local consistency & Bad Global consistency \& texture artifacts \& Janus  & 2mins\\
        Text2Tex ~\cite{chen2023text2tex} & Shape-aware \& Local consistency & Bad Global consistency \& color/pattern shifts \& Janus & 5mins\\ 
        \hline
        \textbf{\MethodName} & Shape-aware \&  Local/Global consistency \& Very fast & Disharmonious patterns \& Janus & \mytilde 4.5secs\\% &  \\
      \bottomrule
    \end{tabular}}
    \label{tab:texture_quantitative}
    \vspace{-0.25cm}
\end{table*}

\subsection*{Mesh Deformer Details}
\begin{itemize}
    \item \textbf{Alignment}: Using the nvdiffmodelling library the base mesh is aligned to the target mesh using the unit-size function that moves/rescales the input to match bounding boxes.
    \item \textbf{Deformation}: We use the same formulation with the Neural Jacobian Fields(NJF) as described in Sec.3.1 of their paper and Sec.3 of TextDeformer(TD). Once the deformation map is obtained using Eq.(1) we obtain the updated vertices of the input mesh. While NJF could deform a garment to have a different pose it’s not possible to change its style (t-shirt->sea-creature-armor) because of the supervision signals. Our goal was not to do mesh-registration but instead stylize input base garment templates via deformation. Hence NJF was chosen for its versatility across heterogeneous mesh collections, because it’s triangulation-agnostic and it provided a flexible and easy-to-use framework to accomplish our goal in a fast plug-n-play manner. 
    \item \textbf{Losses}: Our goal was to deform the base mesh enough to match the pseudo-ground-truth mesh extracted from the image but not fully since if that was the case we’d end up with a watertight mesh unable to be fit to parametric bodies and simulated. Hence we opted for point-to-point meshes (instead of point-to-mesh) for the 3D supervision as well as embedding and image-based losses in the 2D space. 
\end{itemize}

\subsection*{Texture Details}

This approach prioritized filling in the large areas first before moving on to smaller and more occluded regions.

\noindent\textbf{Texture Comparisons}: In Table~\ref{tab:texture_quantitative} we provide a comparison between the pros and cons of recent texture estimation approaches. Our approach is significantly faster compared to past works due to key optimizations described in the main paper while maintaining good local and global consistency. Similar to past works one can notice the Janus problem appearing in texture maps, which can be handled by training a multiview generation diffusion model with more explicit camera pose injection in the future works. 

\begin{figure}[t]
    \centering
    \includegraphics[width=0.99\linewidth]{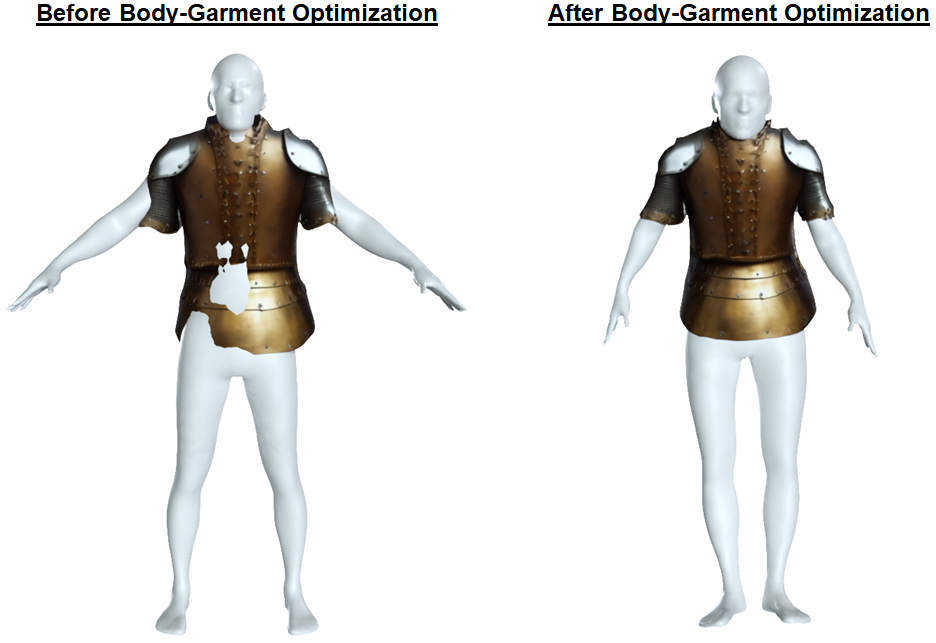}
    \caption{\textbf{Fitting a parametric body to a generated garment}: We start with the generated textured 3D garment (in this case a medieval armor) and a parametric body in its canonical pose (left). After the body-garment optimization process the body pose and shape parameters are optimized such that the generated garment can fit in the body accurately without penetrations.}
    \label{fig:fitting}
\end{figure}

\noindent\textbf{Quantitative Comparisons - Details}: In terms of garment base meshes we utilized the publicly available dataset provided by DiffAvatar which comprises 6 template geometries. For each one of the 6 garments we provide 4 different image inputs (2 real and 2 AI-generated) and to quantitatively evaluate the different approaches we render the untextured outputs of all methods from 36 views. No prompts are used during the quantitative evaluation procedure.

\subsection*{Garment Fitting to Parametric Bodies}\label{ssec:fitting}
While the aforementioned supervisions and regularizations aim to ensure that the quality of the generated garments will be satisfactory for simulation. The produced garment will still need to be scaled, positioned and oriented to fit the parametric body~\cite{smpl_loper} on which it is to be draped and simulated. 
To accomplish this task, we run an optimization procedure during which the generated garment remains fixed in the generated pose and the pose and shape of the parametric body are transformed such that the garment can accurately fit the body.
This optimization process shown in in Fig.~\ref{fig:fitting}, starts with a rigid transformation and scaling of the body and continues with an optimization of the body pose and shape using the Chamfer distance loss.

\noindent\textbf{Rigid Transformation and Scaling}:
The optimization is initialized by applying a rigid transformation (rotation, translation and scaling) of the parametric body model. This step roughly aligns the body with the garment.
\begin{figure}[t]
    \centering
    \includegraphics[width=0.99\linewidth]{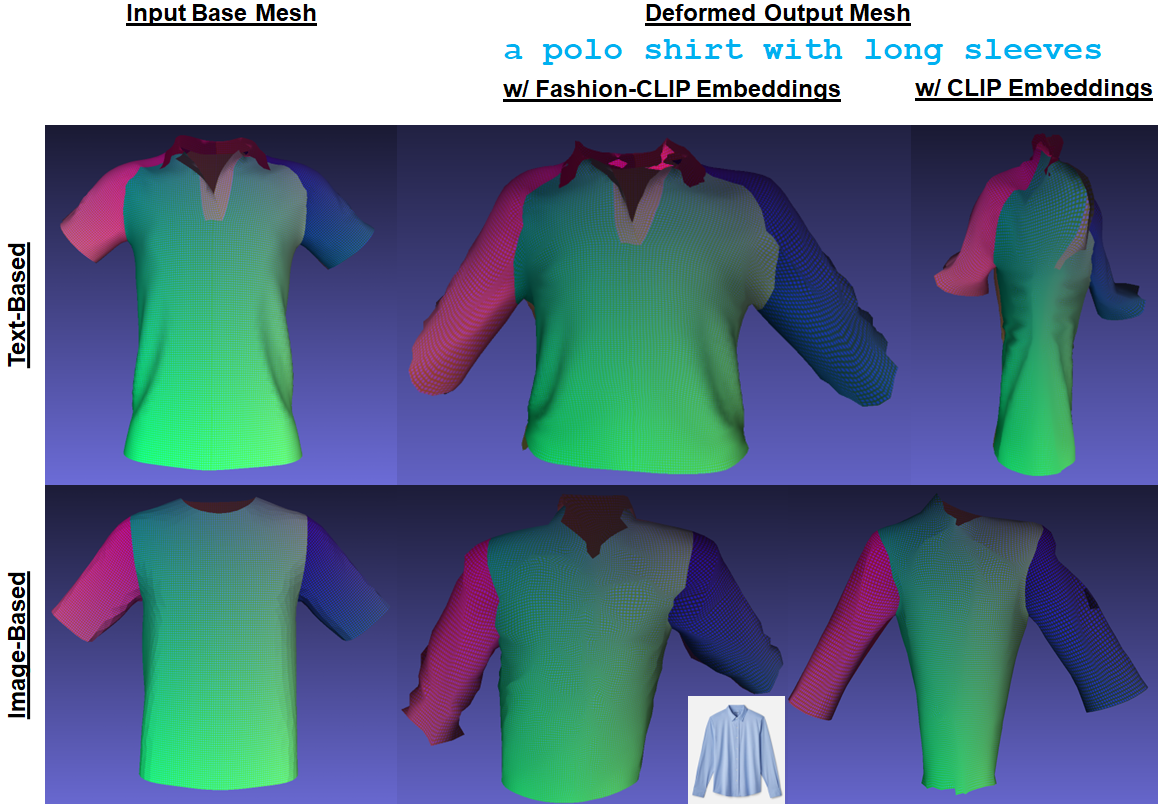}
    \caption{\textbf{Impact of the pre-trained CLIP on garment data}: We disable all other supervisions and explore the impact of a pre-trained CLIP model on fashion data versus using the regular model to enforce embedding supervisions. We observe that regular CLIP embeddings result in distorted and unusable geometries regardless of whether the input is a text prompt or an image.}
    \label{fig:fclip}
\end{figure}

\noindent\textbf{Pose and Shape Optimization}:
Subsequently, the body pose and shape parameters are optimized to minimize the Chamfer distance between the body mesh and the garment mesh. 

\noindent\textbf{Collision Handling}:
After body model optimization, an additional step is performed to resolve potential body-cloth collisions. This is achieved by minimizing an interpenetration loss that penalizes any intersections between the body mesh and the garment mesh.% ensuring a realistic and collision-free fit. 
By combining these steps, the garment mesh can be accurately fit to the parametric body model, enabling realistic draping and physics-based cloth simulations for downstream applications.

\begin{figure}[t]
    \centering
    \includegraphics[width=0.99\linewidth]{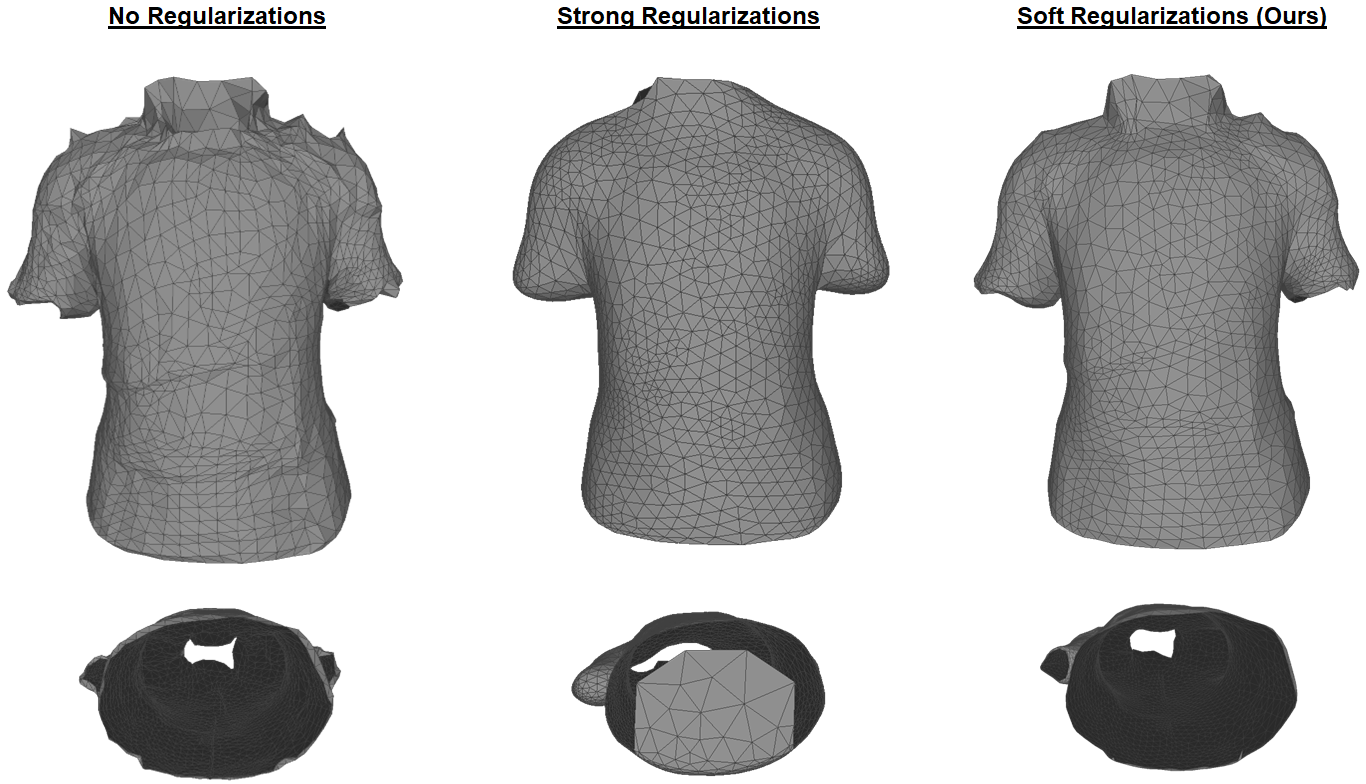}
    \caption{\textbf{Impact of regularizations on the final armor geometry}: Enforcing no regularizations (Laplacian smoothing, penalization of small triangles etc.) on the output mesh results in a crisp output armor mesh with arm/body holes but its quality is not at the level required to perform physics-based simulation. On the other hand, enforcing strong regularizations results in overly smoothed meshes with closed holes. Our output strikes a good balance between capturing those fine-level details that make an armor geometry look like one yet making it suitable for downstream tasks.}
    \label{fig:regularize}
\end{figure}

\subsection*{Additional Ablation Studies}
\noindent\textbf{Supervisions}: When it comes to supervisions we observed that: a) utilizing regular CLIP embeddings provides minimal supervision guidance when it comes to garments and results in poorly deformed meshes which is why we opted for a garment fine-tuned model as shown in Fig.~\ref{fig:fclip}, b) explicitly enforcing multi-view consistency losses is not necessary as 3D supervisions can provide better guidance, and iii) there is a trade-off between allowing for heavy garment stylizations/deformations and maintaining a good mesh quality that can be used later on as shown in Fig.~\ref{fig:regularize}. 
Thus we propose to use a combination of 3D supervisions to guide the deformation process to obtain an accurate 3D shape along with 2D and embedding supervisions to obtain the fine-level details of the garment that the 3D pseudo ground-truth might fail to capture. We train for \mytilde 1000 iterations with the weights of each loss described in Eq.~(6) as follows: \(w_{CD}=20, w_{Lap}=1, w_{triag}=1, w_{2D}=2, w_{E}=4\) with the weight of \(W_{CD}\) gradually decreasing after the first 500 iterations once we have obtained a fairly accurate pose and shape of the garment to allow for the remaining of the supervisions to distill the fine-level garment details. Note that if we were to enforce strong 3D supervisions we would end up with deformed garments that would have no holes for the body, arms and head. 

\begin{figure}[t]
    \centering
    \includegraphics[width=0.99\linewidth]{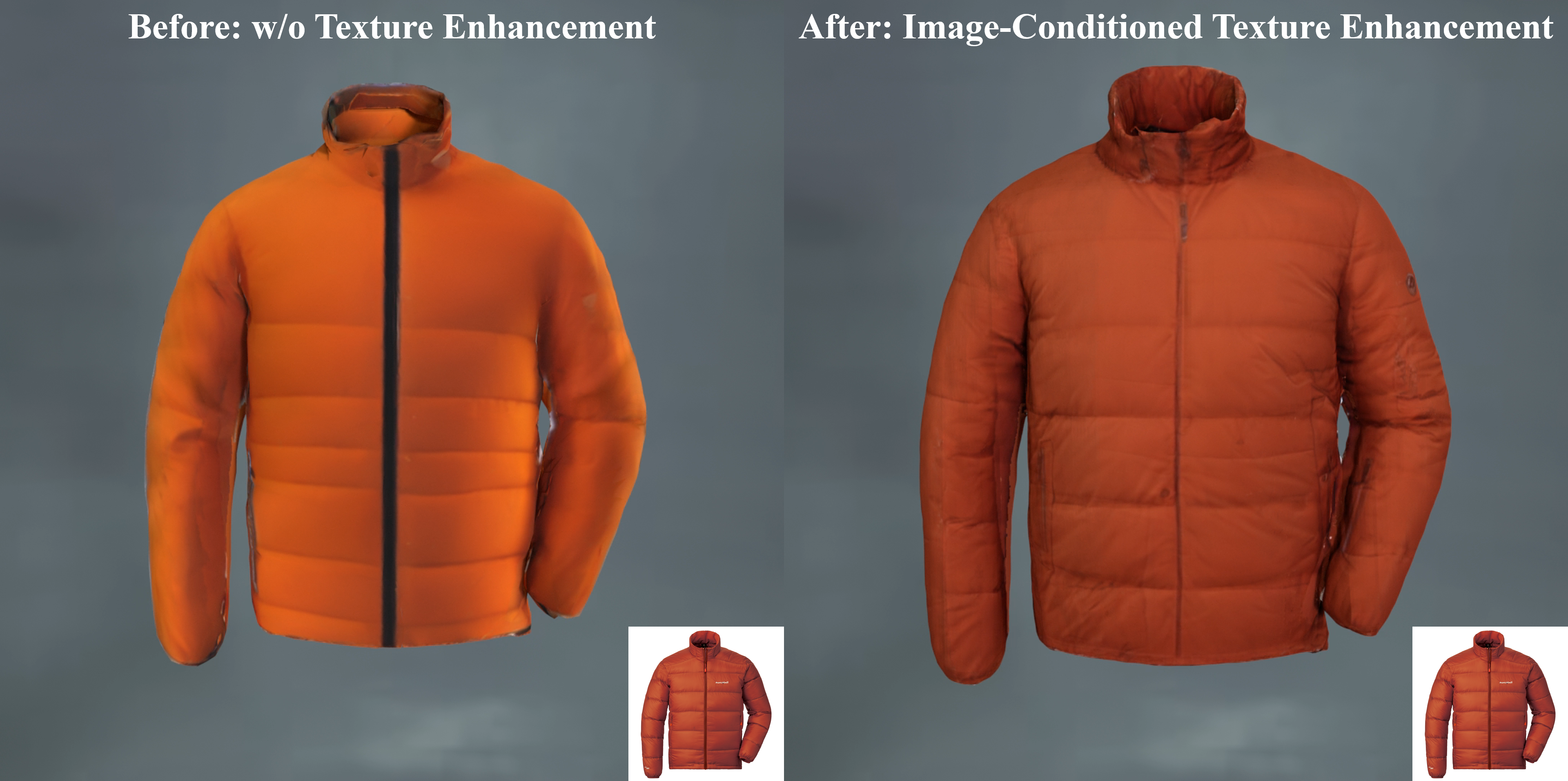}
    \caption{\textbf{Impact of Texture Module}: given the left image as a condition, the texture enhancement module enriches the details and enhances the overall image quality by effectively utilizing the powerful 2D priors.}%\xiaoyu{Can you add one sentence here}
    \label{fig:textures}
\end{figure}

\begin{figure}[t]
    \centering
    \includegraphics[width=0.99\linewidth]{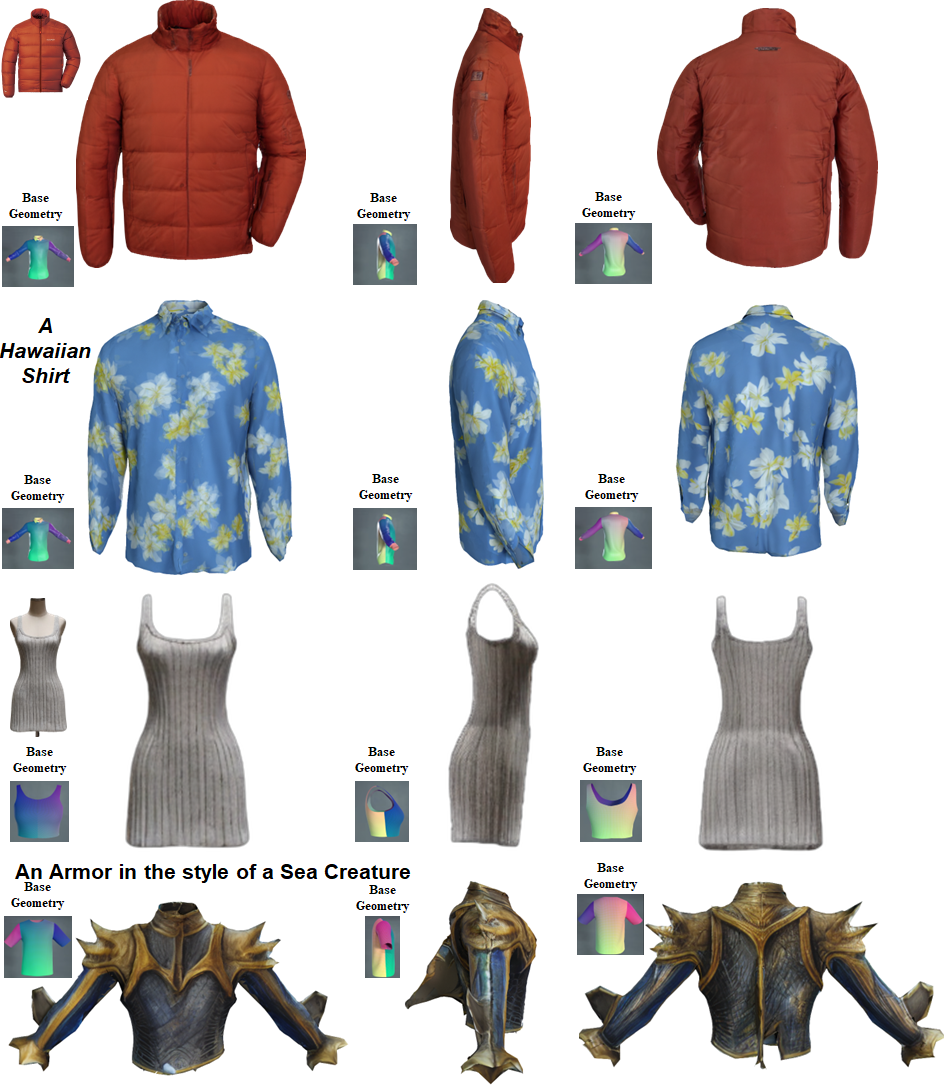}
    \caption{\textbf{3D Garment Generation}: Given an image (\(1^{st}\) row) or a text prompt (\(2^{nd}\) row) as guidance and a base geometry mesh (bottom left inset) that can be far from the target we generate high-quality textured 3D geometries of both real as well as fantastical garments.} 
    \label{fig:qualitative}
    \vspace{-0.2cm}
\end{figure}

\begin{figure*}[t]
    \centering
    \includegraphics[width=0.99\linewidth]{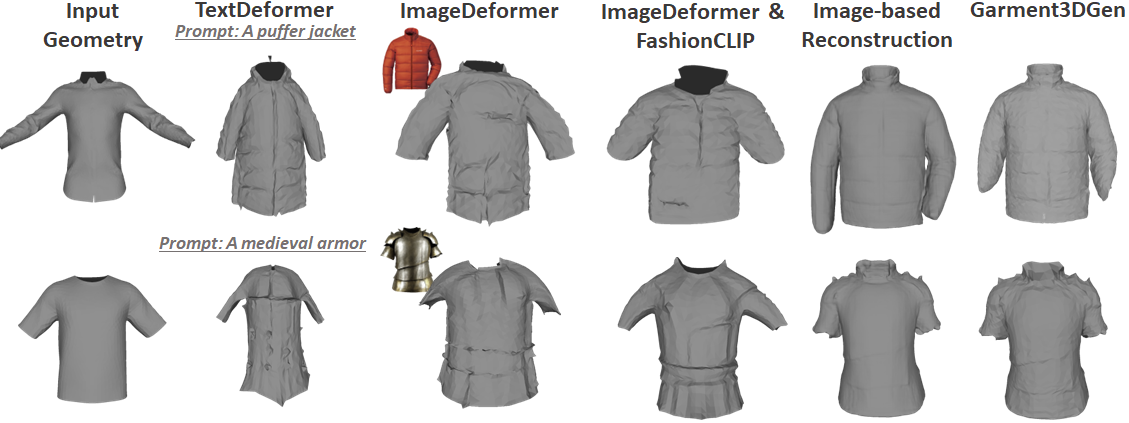}
    \caption{\textbf{Ablation Study}: Starting from a base input mesh we showcase that our key contributions result in deformed geometries that capture the input image guidance, comprise fine-level garment details and are suitable for our downstream tasks.}
    \label{fig:ablation}
    \vspace{-0.25cm}
\end{figure*}

\begin{figure*}[t]
    \centering
    \includegraphics[width=0.99\linewidth]{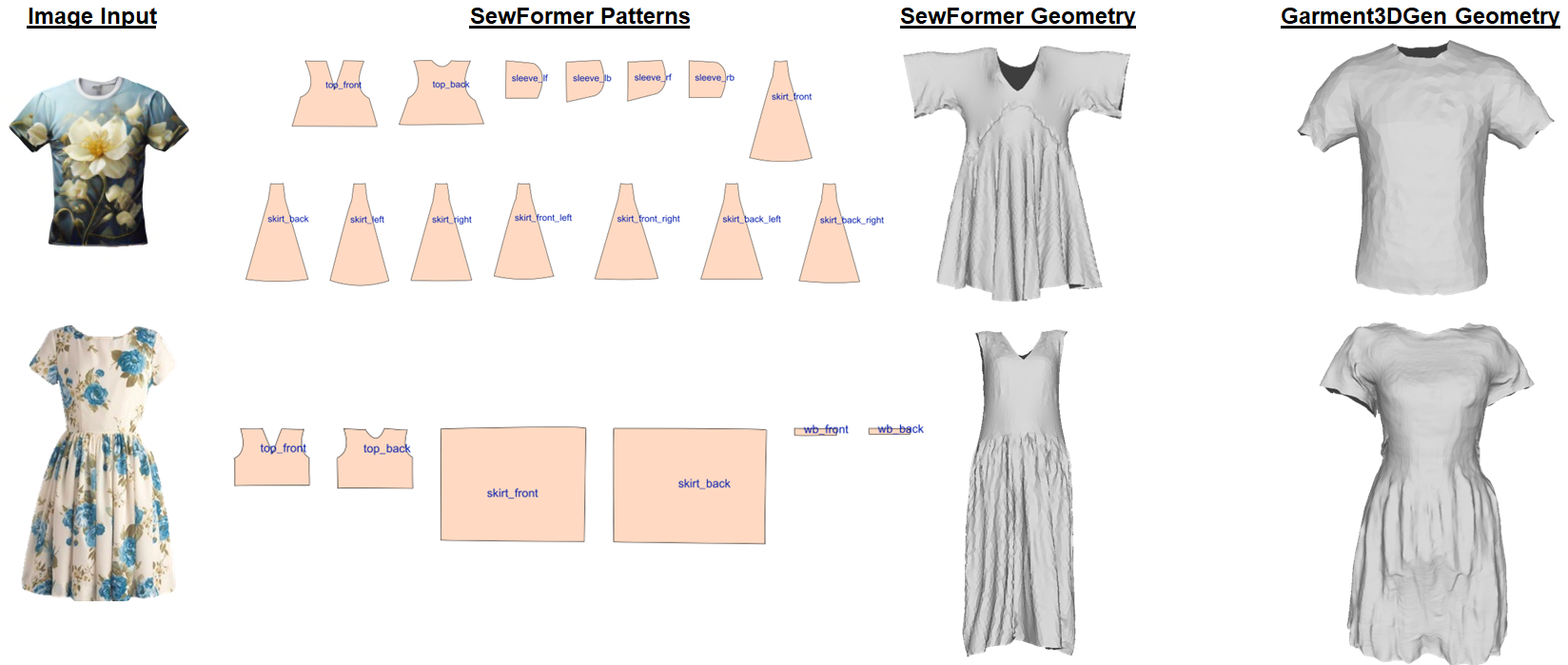}
    \caption{\textbf{Qualitative Comparisons}: SewFormer~\cite{liu2023towards} produces 2D garment patterns from a single RGB image input. We further process these panels by virtually sewing them together at the seams to create assembled 3D garments, which enables us to perform a qualitative comparison. We showcase two garment examples and demonstrate that Garment3DGen produces results that are more faithful to the image input with a less complex pipeline.}
    \label{fig:sewformer}
\end{figure*}

\noindent\textbf{Adding Components one at a time}: As described in the main paper we conducted an ablation study depicted in Fig.~\ref{fig:ablation}. We start with the off-the-shelf TextDeformer which takes a text prompt and a base-mesh and deforms this to match the target text. Text prompts are not ideal to capture the fine-level details of a garment as there can be many ``medieval armors''. In addition, a pretrained CLIP model is not capable of capturing the subtle differences between a ``jacket'' and a ``puffer jacket''. 
To overcome this limitation we adapt TextDeformer to take image inputs as guidance (ImageDeformer) and observe that the deformed geometries are improved. Nonetheless, they still fail to capture the details of the image. By swapping the original CLIP model and introducing a model that is fine-tuned on fashion data we observe that details are better preserved across garments. 
Noting that image-based reconstruction methods can accurately capture geometry but produce coarse and watertight meshes that are unsuitable for subsequent tasks, we utilize these meshes as pseudo ground-truth for our proposed approach.
Our \MethodName results in garments that faithfully follow the image guidance while containing the wrinkles and fine details. 
However the quality of the output geometry is not always ideal for physics-based downstream tasks because they produce poorly conditioned triangles which will result in instabilities when simulated in addition to poorly tessellated geometry which will result in unnatural fabric behavior. Because of this, we introduced additional 3D supervisions that preserve a better mesh quality.

\noindent\textbf{Texture Module}: The impact of the texture enhancement module is shown in Fig.~\ref{fig:textures}. The textures directly synthesized by 3D generation models tend to be low-resolution, smooth and over-simplified, which is due to the scarcity of high quality 3D training data. Thus, the texture enhancement module aims to effectively utilize the 2D priors learned from the large high-quality image dataset. After our image-conditioned image enhancement, we bring back vivid details to the texture, improving the perceptual quality. 

\subsection*{Additional Results}
\noindent\textbf{Comparisons with SewFormer}: In Fig.~\ref{fig:sewformer} we provide a qualitative comparisons against SewFormer~\cite{liu2023towards} which predicts garment patterns from a single RGB image.To facilitate a meaningful comparison, we use a physics-based cloth simulator to sew the generated panels together, creating an assembled 3D garment. In the top row T-shirt example, SewFormer incorrectly generates pattern pieces for a dress instead of the requested T-shirt. After assembly, it's clear that the predicted garment resembles a dress rather than the intended T-shirt. The second example shows that while SewFormer works reasonably well for certain garments, it fails to fully match the guidance, missing sleeves in this case. Both examples demonstrate that SewFormer produces results that don't accurately match the image guidance, and its method for obtaining 3D assembled garments is more complex, requiring specialized software to sew individual pattern pieces together. Additionally, SewFormer is limited to producing V-neck designs, whereas our method correctly follows the visual guidance.

\begin{figure*}[t]
    \centering
    \includegraphics[width=0.99\linewidth]{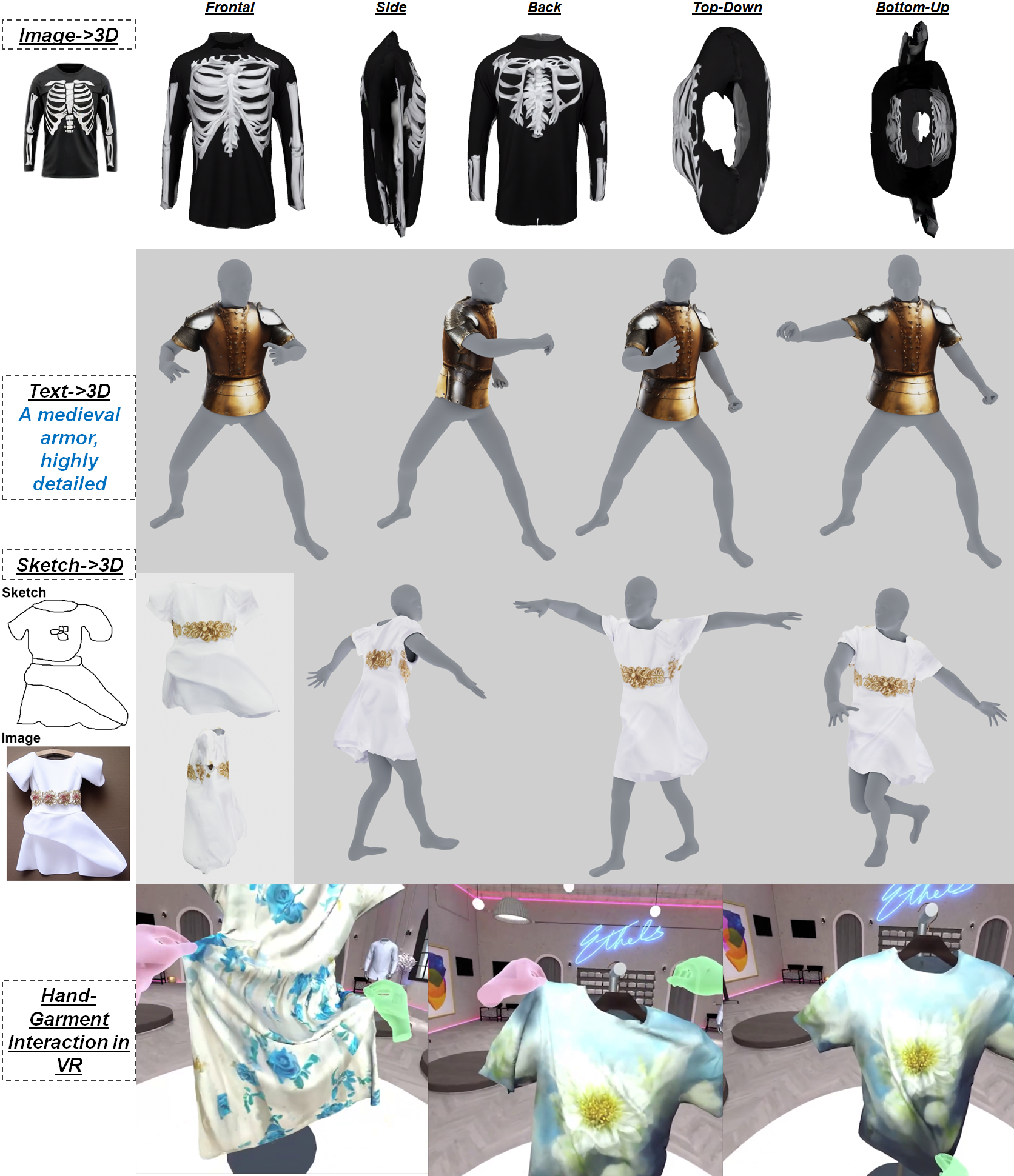}
    \caption{\textbf{Garment3DGen Applications}: We showcase various applications of our proposed approach.}
    \label{fig:Qual_FullPage2}
\end{figure*}

\noindent\textbf{Additional Qualitative Comparisons}: In Fig.~\ref{fig:qualitative} we provide multi-view renders of our 3D textured garments that we generate from text prompts and an image guidance. From these results we gain the following insights: a) \MethodName works just as well with fantastical garments (armors or dresses) that are outside the regular garment distribution, b) our texture estimation module results into high-quality textures that closely match the input text prompt and c) our output geometry does not have to be similar to the input base mesh. Finally in Fig.~\ref{fig:Qual_FullPage2} we showcase the plethora of applications that Garment3DGen has ranging from text/image/sketch to simulation-ready 3D garments to hand-garment interaction in a VR environment using the on-device hand-tracking. 

\end{document}